\documentclass[10pt,twocolumn,letterpaper]{article}

\usepackage[pagenumbers]{cvpr} 

\usepackage{colortbl}
\usepackage{graphicx}
\usepackage{subcaption}
\usepackage{xspace}
\usepackage{multirow}
\usepackage{makecell}
\usepackage{algorithm}
\newcommand{\best}{\cellcolor{tablered}}
\newcommand{\sbest}{\cellcolor{orange}}
\newcommand{\tbest}{\cellcolor{yellow}}

\newcommand{\bD}{\mathbf{D}}

\newcommand{\bn}{\mathbf{n}}\newcommand{\bN}{\mathbf{N}}

\newcommand{\bx}{\mathbf{x}}

\newcommand{\bmu}{\boldsymbol{\mu}}

\newcommand{\bsigma}{\boldsymbol{\sigma}}\newcommand{\bSigma}{\boldsymbol{\Sigma}}

\newcommand{\nR}{\mathbb{R}}

\newcommand{\cG}{\mathcal{G}}

\newcommand{\cL}{\mathcal{L}}

\newcommand{\boldparagraph}[1]{\vspace{0.1cm}\noindent{\bf #1.}}

\definecolor{yellow}{rgb}{1, 1, 0.7}
\definecolor{orange}{rgb}{1, 0.85, 0.7}
\definecolor{tablered}{rgb}{1, 0.7, 0.7}
\definecolor{red}{rgb}{1, 0, 0}

\definecolor{wincolor}{rgb}{0.85, 0.0, 0.0}

\definecolor{darkyellow}{rgb}{0.8, 0.8, 0.5}
\definecolor{darkred}{rgb}{0.7, 0.3, 0.3}
\definecolor{darkgreen}{rgb}{0.3, 0.7, 0.3}
\definecolor{blue}{rgb}{0.125, 0.469, 0.703}
\definecolor{lightskyblue}{rgb}{0.4706, 0.1922, 0.0000}
\definecolor{green}{rgb}{0, 1.0, 0}
\definecolor{pink}{rgb}{1, 0.4, 0.7}

\definecolor{cvprblue}{rgb}{0.21,0.49,0.74}
\usepackage[pagebackref,breaklinks,colorlinks,allcolors=cvprblue]{hyperref}

\def\paperID{7840} 
\def\confName{CVPR}
\def\confYear{2025}

\title{
SolidGS: Consolidating Gaussian Surfel Splatting for \\ Sparse-View Surface Reconstruction
}

\author{Zhuowen Shen$^{1,2}$$^{*}$ \quad Yuan Liu$^{3,4}$ \quad Zhang Chen$^{2}$ \quad Zhong Li$^{2}$ \quad Jiepeng Wang$^{5}$ \quad Yongqing Liang$^{1}$\\ \quad Zhengming Yu$^{1}$ \quad Jingdong Zhang$^{1}$ \quad Yi Xu$^{2}$ \quad Scott Schaefer$^{1}$ \quad  Xin Li$^{1}$ \quad Wenping Wang$^{1}$ \\[0.3em]
$^{1}$Texas A\&M University \quad $^{2}$OPPO US Research Center \quad $^{3}$Nanyang Technological University \\
$^{4}$Hong Kong University of Science and Technology \quad $^{5}$The University of Hong Kong}

\begin{document}

\twocolumn[{
\renewcommand\twocolumn[1][]{#1}
\maketitle
    \vspace*{-7ex}
    \begin{center}
    \includegraphics[width=0.95\textwidth]{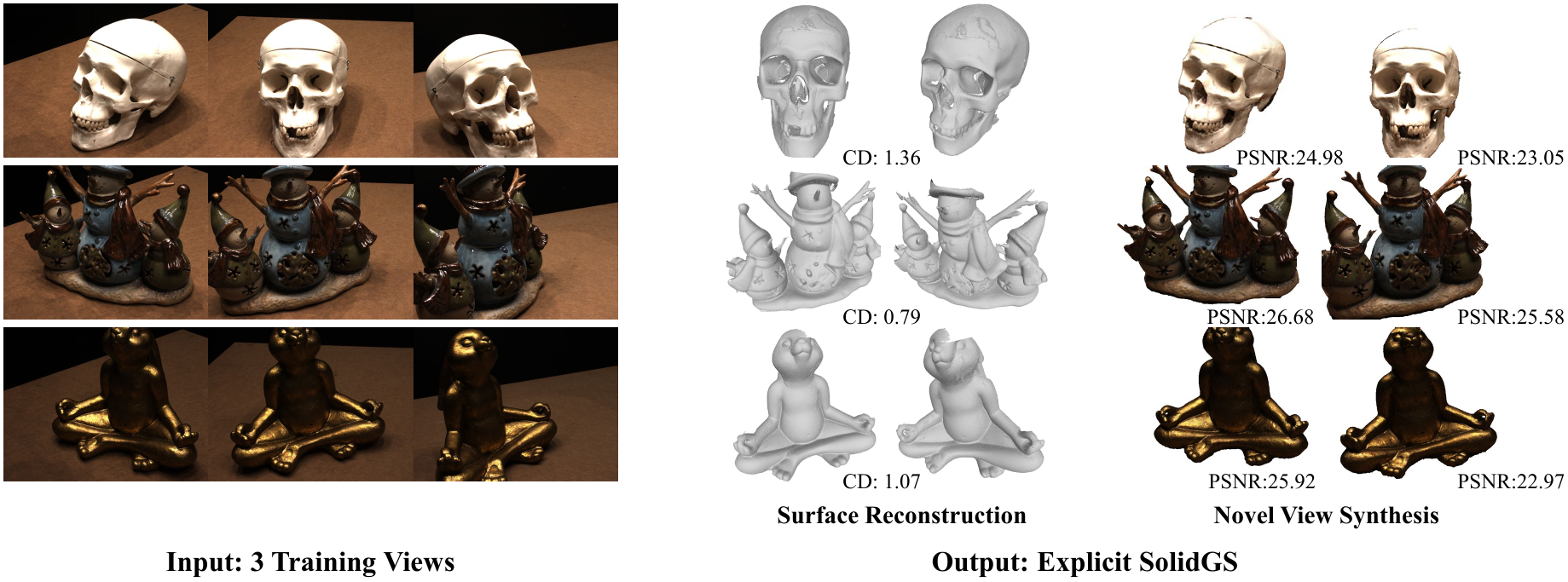}
    \end{center}  
    \vspace*{-15px}
    \captionof{figure}{\textbf{Overview of SolidGS.} We present SolidGS, which reconstructs a consolidated Gaussian field from sparse inputs. Given only three input views, our approach enables high-precision and detailed mesh extraction, and high-quality novel view synthesis, achieved within just three minutes.
    }
    \label{fig:teaser}
}
]
\maketitle

\begin{abstract}
Gaussian splatting has achieved impressive improvements for both novel-view synthesis and surface reconstruction from multi-view images. However, current methods still struggle to reconstruct high-quality surfaces from only sparse view input images using Gaussian splatting. In this paper, we propose a novel method called SolidGS to address this problem. We observed that the reconstructed geometry can be severely inconsistent across multi-views, due to the property of Gaussian function in geometry rendering. This motivates us to consolidate all Gaussians by adopting a more solid kernel function, which effectively improves the surface reconstruction quality. With the additional help of geometrical regularization and monocular normal estimation, our method achieves superior performance on the sparse view surface reconstruction than all the Gaussian splatting methods and neural field methods on the widely used DTU, Tanks-and-Temples, and LLFF datasets.

\end{abstract}

{\let\thefootnote\relax\footnote{{ \scriptsize
{$^{*}$}Work partially done during his internship at OPPO US Research Center.
 }}}

\vspace{-4mm}
\section{Introduction}
\vspace{-1mm}

\begin{figure*}[t]
  \centering
   \vspace{-5mm}
   \includegraphics[width=0.95\linewidth]{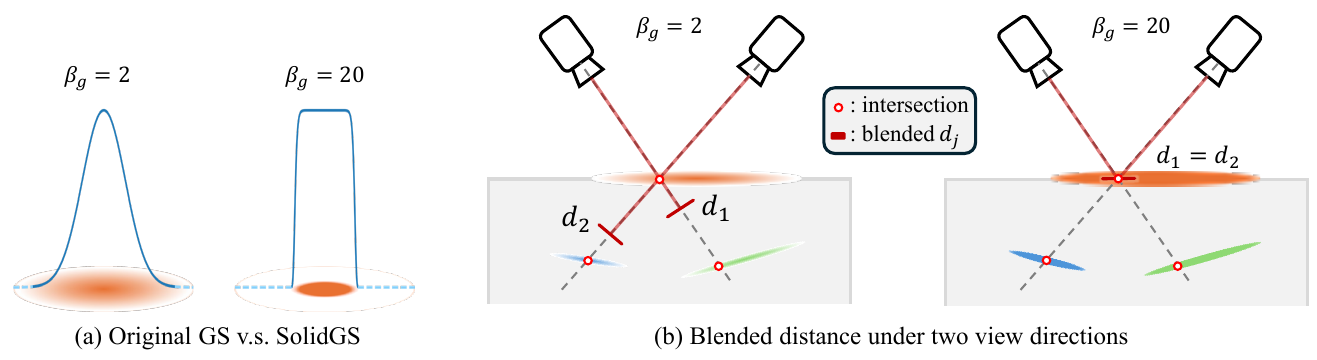}
   \vspace{-3mm}
   \caption{\textbf{Illustration of SolidGS.} (a) The Gausian functions \(\cG_i\) of the original 3DGS and SolidGS in \cref{eq:solidgaussian}. 
   (b) Visualization of ray-plane intersections and blended depths \(d_j, j \in \{1,2\}\) for two types of Gaussians. In the original Gaussian, as rays deviate from the Gaussian center, the depth is blended with the background Gaussians, leading to inconsistent multi-view geometry in 3D space. Meanwhile, our SolidGS is opaque in most of its effective areas, giving out consistent geometry regardless of view directions. } 
   \label{fig:gef}
   \vspace{-3mm}
\end{figure*}

\label{sec:intro}



Accurately reconstructing the geometry while maintaining photo-realistic novel view synthesis results has always been a popular topic in the field of 3D computer vision~\cite{schoenberger2016mvs}. Previously, many works~\cite{wang2021neus, li2023neuralangelo, yariv2021volume} built on Neural Radience Field (NeRF)~\cite{mildenhall2021nerf} incorporating Signed Distance Function (SDF) have achieved remarkable reconstruction results. However, these neural representation methods usually suffer from hours of training time, and rendering an image is usually inefficient for real-time applications such as AR and VR. 

Recently, 3D Gaussian Splatting (3DGS)~\cite{kerbl20233d} adopts an explicit point-based representation and achieves minutes-long training with comparable NVS quality. The succeeding works~\cite{huang20242d,chen2024pgsr,yu2024gaussian} extend 3DGS to reconstruct high-quality surfaces, achieving comparable results to NeRF-based methods. The key idea of these explicit methods~\cite{huang20242d,xu2024mvpgs} is to squash Gaussian primitives to approximate surfaces and explicitly calculate the exact intersection points between the camera rays and Gaussian primitives in the rendering. With the exact intersections, these works can render high-quality normal and depth maps from the Gaussians and constrain the surface by applying geometry regularization.
Though the aforementioned works made tremendous progress in multi-view surface reconstruction, these methods usually require hundreds of multi-view images as input. 
In many applications such as AR/VR or robotics, we only have access to sparse views and these methods struggle to reconstruct reasonable surfaces from these sparse view images.

We find that the main reasons for the degraded reconstruction quality are the multi-view depth inconsistency and insufficient supervision from sparse input.
As shown in \cref{fig:gef}, though existing methods try to squash the Gaussian primitives into planes, a strong depth inconsistency still exists. Due to the property of the Gaussian function during alpha blending, the opacity of a Gaussian primitive unavoidably decreases as deviating from its center, which causes different rendered depth values for two different rays that intersect at the same point on the surface. Consequently, the inconsistent depth causes noisy reconstructed surfaces in existing methods~\cite{huang20242d,chen2024pgsr}. Meanwhile, the sparse view inputs fail to provide enough constraints for the Gaussian opacity to converge to solid surfaces, which further degrades the reconstruction quality.

In this paper, we propose a novel method, SolidGS, to tackle the above challenges for sparse-view surface reconstruction. The key ideas of our method are consolidating the Gaussians and introducing additional geometry constraints. These constraints consist of self-supervision geometry loss from virtual views and monocular normal regularization. Both the consolidation of Gaussians and the proposed constraints greatly improve the surface reconstruction quality within the sparse view setting.

As shown in Fig.~\ref{fig:gef} (b), consolidating the Gaussian primitives into our SolidGS prevents camera rays from penetrating the Gaussian primitives, thereby substantially reducing the rendered depth inconsistency and enhancing reconstruction quality. 
A naive way for such consolidation is to directly apply a constant opacity value for the primitive to replace the Gaussian function. However, this method poses challenges for stable optimization, as it causes differentiability issues over the primitives~\cite{liu2019soft, ravi2020accelerating}.
To promote stable optimization while consolidating Gaussian primitives, we get inspiration from GES~\cite{hamdi2024ges} to adopt a trainable exponential factor within the Gaussian function and encourage the exponential factor to converge to a large value by sharing the factor for all Gaussian primitives. This strategy enables our representation to be a vanilla Gaussian representation in the beginning while gradually stabilizing into solid primitives throughout the optimization process.

In addition to consolidating Gaussian primitives, we incorporate several novel constraints to guide our optimization process toward accurate surface reconstruction. Specifically, we introduce additional self-supervised geometry regularization from virtual cameras. We also estimate normal maps from input views using monocular normal estimators~\cite{hu2024metric3d}. These virtual view regularizations and normal maps stabilize the optimization and lead to accurate surface reconstruction.

We have conducted extensive experiments on the DTU~\cite{jensen2014large}, Tanks and Temples~\cite{knapitsch2017tanks}, and LLFF~\cite{mildenhall2019local} dataset. The results show that our method reconstructs high-quality surfaces even only given 3 input RGB images in 3 minutes, which greatly improves the reconstruction quality than Gaussian splatting-based methods~\cite{xu2024mvpgs, li2024dngaussian} while being much more efficient than previous neural SDF-based baselines~\cite{huang2024neusurf, long2022sparseneus}.

In summary, our major contributions are threefold:
\begin{itemize}
\setlength{\itemsep}{0pt}
\setlength{\parskip}{2pt}
    \item We propose SolidGS, a novel representation that consolidates the opacity of Gaussians by introducing a shared, learnable solidness factor, enabling multi-view consistent geometry rendering. 
    \item We introduce a new framework with geometric constraints to train our SolidGS representation, which consists of geometric priors and regularizations.
    \item We perform extensive experiments on DTU, Tanks and Temples, and LLFF datasets. Our method outperforms existing state-of-the-art methods on these datasets.
\end{itemize}

\section{Related works}
\label{sec:rw}

\subsection{Neural Volume Rendering}

Neural volume rendering uses neural networks to represent 3D scenes as continuous volumetric data for realistic, view-dependent rendering, which was first introduced in Neural Volume~\cite{lombardi2019neural}. It surpasses traditional methods by producing seamless, photorealistic views and is essential for applications like VR and 3D graphics. 
A pivotal development was the introduction of Neural Radiance Fields (NeRF)~\cite{mildenhall2021nerf}. NeRF employed a differentiable volumetric rendering technique to reconstruct a neural scene representation, achieving impressive photorealistic view synthesis with view-dependent effects. To accelerate its optimization, subsequent research replaces the neural scene representation with explicit or hybrid scene representations, such as voxel grids~\cite{fridovich2022plenoxels,sun2022direct}, low-rank tensors~\cite{chen2022tensorf}, tri-planes~\cite{chan2022efficient}, multi-resolution hash grids~\cite{muller2022instant}, and even point pivoted radiance filed~\cite{xu2022point}. 
Recently, 3D Gaussian Splatting (3DGS)~\cite{kerbl20233d} demonstrates the possibility of 3D Gaussians modeling continuous distributions of color and opacity across space, significantly enhancing rendering speed and reducing memory consumption. More succeeding works~\cite{qu2024disc, liang2024analytic, yu2024mip} continue to refine the visual quality. 
To further exploit the power of Gaussian Splatting, our work focuses on a more challenging but practical sparse-view setting.

\subsection{Novel View Synthesis from Sparse View}

Sparse-view Novel View Synthesis (NVS), which focuses on rendering high-quality images from limited camera perspectives, has garnered extensive research interest in recent years. NeRF-based methods improve sparse-view reconstruction by leveraging additional constraints~\cite{niemeyer2022regnerf, yang2023freenerf, somraj2023vip}, multi-view stereo priors~\cite{chen2021mvsnerf}, and depth priors~\cite{deng2022depth, wang2023sparsenerf, roessle2022dense}. While these methods perform optimization for the individual scene, there are also works that directly reconstruct the scene in a feed-forward manner~\cite{yu2021pixelnerf, suhail2022generalizable, du2023learning}. However, these works still suffer from heavy volume sampling and long training time.
Gaussian-based techniques have emerged as an efficient solution. Many works have engaged in-depth recently~\cite{zhu2025fsgs, bao2024loopsparsegs, paliwal2025coherentgs, xiong2023sparsegs}. DNGaussian~\cite{li2024dngaussian} leveraging depth regularization for efficient and high-quality few-shot NVS. MVPGS~\cite{xu2024mvpgs} extends this approach by integrating multi-view priors to improve geometry consistency across views, thereby enhancing performance in sparse-view settings. Gaussian-based feed-forward methods also achieve superior quality~\cite{chen2025mvsplat, charatan2024pixelsplat}.
Our primary target is to reconstruct multi-view consistent geometry.

\subsection{Surface Reconstruction from Sparse View}
Accurately reconstructing geometry simultaneously during neural volume rendering has gained significant interest recently. NeuS~\cite{wang2021neus} and VolSDF~\cite{yariv2021volume} learn SDF during radiance field training to achieve accurate geometry reconstruction. More works~\cite{fu2022geo, wang2022neuris, liang2023helixsurf} improve the geometry quality further. To achieve accurate geometry reconstruction from sparse view inputs~\cite{ren2023volrecon, wu2023s, long2022sparseneus, huang2024neusurf, younes2024sparsecraft}, SparseNeus~\cite{long2022sparseneus} learns generalizable priors from image features for sparse view reconstruction. Neusurf~\cite{huang2024neusurf} leverages on-surface priors obtained from SfM to achieve faithful surface reconstruction. More recently, SparseCraft~\cite{younes2024sparsecraft} regularized the model with learning-free multi-view stereo (MVS) cues without pretrained priors. Although the reconstruction quality has been improved, these methods are still slow in training due to the property of volume rendering. With the recent development of 3D Gaussian Splatting~\cite{kerbl20233d}, SuGaR~\cite{guedon2024sugar} proposed a method to extract Mesh from 3DGS. 2DGS~\cite{huang20242d} achieves view-consistent geometry by collapsing the 3D volume into a set of 2D oriented planar Gaussian disks. GOF~\cite{yu2024gaussian} establishes a Gaussian opacity field, enabling geometry extraction by directly identifying its level-set. PGSR~\cite{chen2024pgsr} improves the quality further by utilizing the MVS priors. However, these methods require dense multi-view inputs to get accurate reconstructed results. To tackle this, we use solid Gaussian Surfels as a representation to achieve fast and accurate geometry reconstruction from sparse views.

\section{Method}

\begin{figure*}[t]
  \centering
\vspace{-2mm}
   \includegraphics[width=0.95\linewidth]{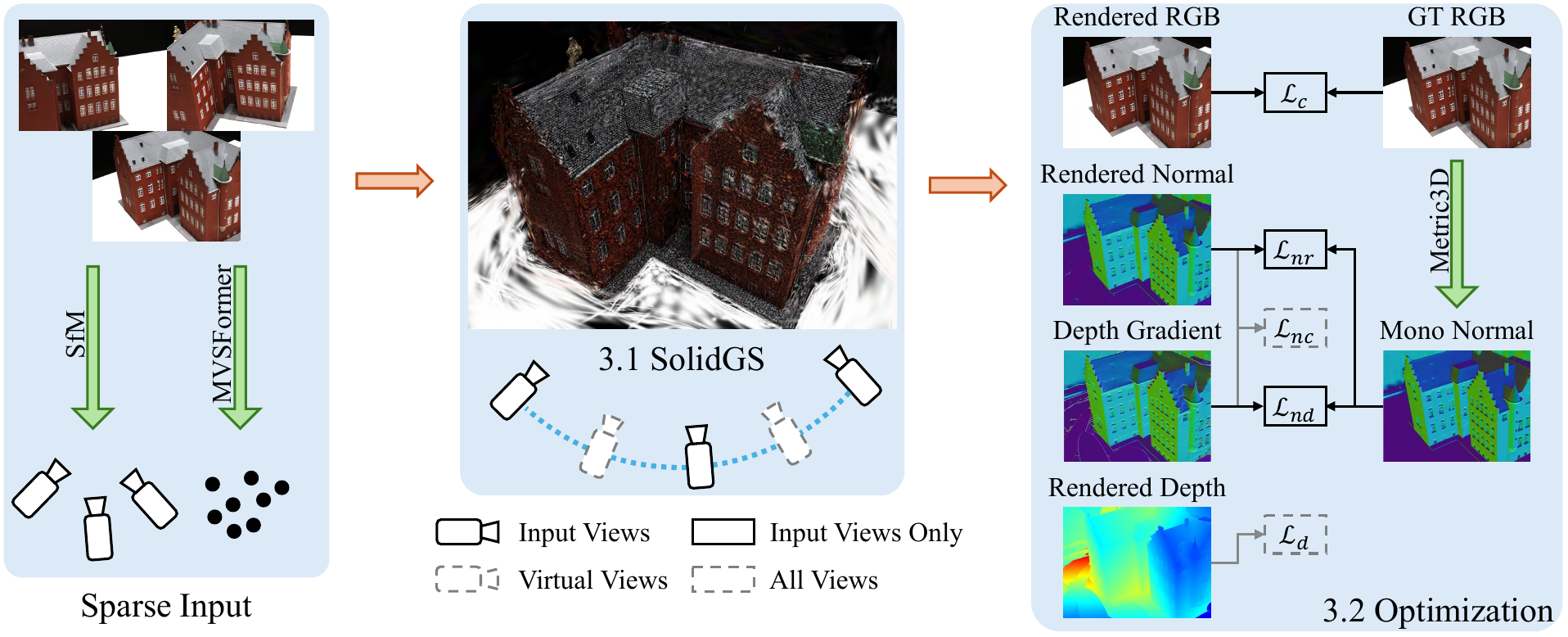}
\vspace{-3mm}
   \caption{\textbf{SolidGS Framework.} With 3 input views, we initialize the camera pose with COLMAP and the point clouds with MVSFormer. Virtual views are generated by linear interpolation between pairs of training views to provide additional geometric regularization. We represent the scene with our SolidGS and gradually enhance the solidity during training. SolidGS are optimized with photometric loss, monocular normal loss, and geometric regularization (normal consistency loss and depth distortion loss).}
   \label{fig:flow}
\vspace{-5mm}
\end{figure*}

Given sparse view inputs (e.g. 3 views), our goal is to reconstruct the surface of the scene with high accuracy and precision while maintaining a satisfactory novel view synthesis result simultaneously. An overview of our method is provided in Fig.~\ref{fig:flow}. In this section, we first review the 3D Gaussian Splatting~\cite{kerbl20233d} in \cref{sec:3dgaussian} and introduce our modified Gaussian representation, SolidGS, in \cref{sec:SolidGS}. To guide the Gaussians to be solid and 3D consistent, we add the geometric regularization in \cref{sec:regularization} alongside geometric clues from the monocular normal estimator in \cref{sec:priors}.

\subsection{SolidGS}

\subsubsection{3D Gaussian Splatting Preliminary}
\label{sec:3dgaussian}

3D Gaussian Splatting~\cite{kerbl20233d} utilizes a set of explicit 3D Gaussians $\{\cG_i\}$ to represent a 3D scene. Each Gaussian is parameterized by an opacity ($\rho_i$), a center location ($\bmu_i \in \nR^{3}$), a color ($c_i \in \mathbb{R}^3$), a rotation ($r_i \in \mathbb{R}^4$) in quaternion form, and a scale vector ($s_i \in \mathbb{R}^3$).  In the world coordinate, the Gaussian distribution is defined as:
\begin{equation}
\label{eq:gaussian}
\cG_i(\bx) = \exp \left\{-\frac{1}{2}\left((\bx-\bmu)^{\top} \boldsymbol{\bSigma}^{-1}(\bx-\bmu)\right)\right\}, 
\end{equation}
\noindent where \(\Sigma_i \in \mathbb{R}^{3 \times 3}\) is the corresponding 3D covariance matrix. The covariance matrix \(\Sigma_i\) can be factorized by $\Sigma_i = R_i S_i S_i^\top R_i^\top$ into a scaling matrix \(S_i \in \mathbb{R}^{3 \times 3}\) and a rotation matrix \(R_i \in \mathbb{R}^{3 \times 3}\).

3DGS enables fast rendering by \(\alpha\)-blending. The color \(C \in \mathbb{R}^3\) of a pixel \(x\) can be obtained through \(\alpha\)-blending:




\vspace{-3mm}
\begin{equation}
C = \sum_{i \in N} T_i \alpha_i c_i, \quad T_i = \prod_{j=1}^{i-1} (1 - \alpha_j),
\end{equation}

\noindent where \(\alpha_i = \rho_i \cG_i(x)\) is the blending weight. \(T_i\) is the cumulative transmittance, and \(N\) is the number of Gaussians that the ray passes through.

For geometry rendering, we use the flattened 3DGS in PGSR~\cite{chen2024pgsr}, and calculate the distance from the plane to the camera center:

\vspace{-4mm}
\begin{equation}
d_i = (\bmu_i - T_c) \cdot n_i,
\end{equation}

\noindent where \(T_c\) is the camera center and $n_i$ is the normal direction corresponding to the minimum scale factor of the Gaussian.

Normal and distance maps are rendered with \(\alpha\)-blending:
\begin{equation}
\bN=\sum_{i \in N} T_i \alpha_i n_i, \quad  D = \sum_{i \in N} T_i \alpha_i d_i.
\end{equation}

Then, the depth map is acquired by intersecting rays with the plane:
\begin{equation}
\bD(p) = \frac{D}{\bN(p) K^{-1} [p, 1]},
\end{equation}

\noindent where \(p = [u, v]^\top\) is the 2D position on the image plane and \(K\) is the intrinsic matrix of the camera.

\subsubsection{Solid Gaussian Representation}
\begin{figure*}[t]
  \centering
   \vspace{-1mm}
   \includegraphics[width=0.95\linewidth]{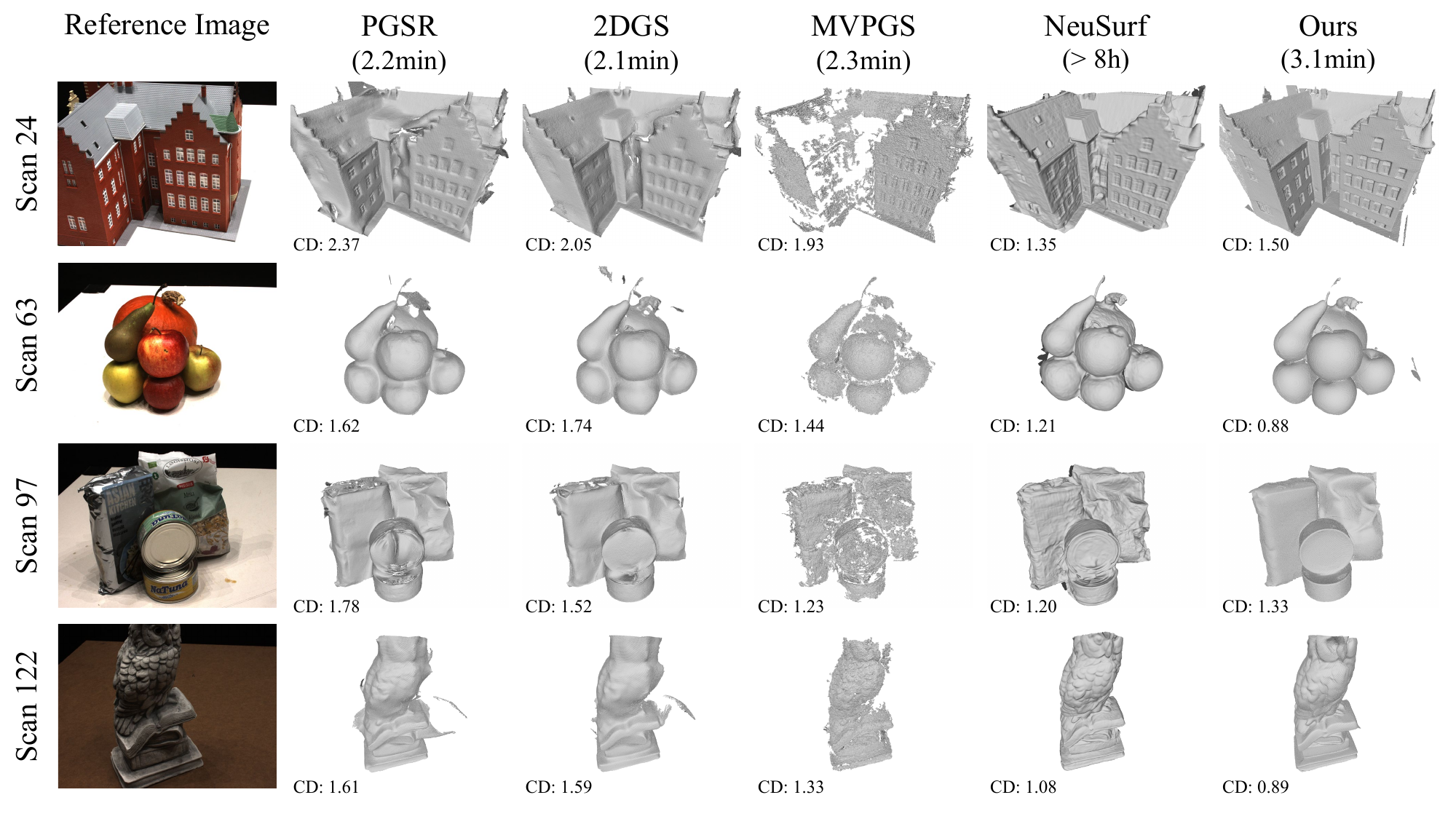}
   \vspace{-3mm}
   \caption{\textbf{Qualitative Mesh Results on DTU Dataset.} We show the reconstructed meshes with the closest input view for reference. Meshes are fused using the TSDF + Marching Cube method for explicit methods including PGSR~\cite{chen2024pgsr}, 2DGS~\cite{huang20242d}, MVPGS~\cite{xu2024mvpgs}, and our method. NeuSurf~\cite{huang2024neusurf} maintains an SDF, from which the mesh is extracted using the Marching Cube.}
   \label{fig:dtu_mesh}
\end{figure*}

\label{sec:SolidGS}
In the original Gaussian splitting, an outstanding issue for geometry rendering is the non-solid nature of Gaussian primitives. As shown in Fig.~\ref{fig:gef} (a), the alpha value of the intersection point decreases quickly as it deviates from the Gaussian center. Therefore, during \(\alpha\)-blending, rendered depth could be inconsistent for two rays intersecting the same 3D Gaussian, as shown in Fig.~\ref{fig:gef} (b). This inconsistency would introduce extra difficulties during optimization and lead to degraded surface reconstruction quality.


To fix this issue, we need to increase the solidity of Gaussians. We substitute the Gaussian distribution in \cref{eq:gaussian} with the generalized exponential Gaussian distribution:

\vspace{-5mm}
\begin{equation}
\label{eq:solidgaussian}
    \cG_i(\bx) = \exp \left\{-\frac{1}{2}\left((\bx-\bmu)^{\top} \boldsymbol{\bSigma}^{-1}(\bx-\bmu)\right)^{\frac{\beta_i}{2}}\right\},
\end{equation}
\vspace{-2mm}

\noindent which was first introduced in GES \cite{hamdi2024ges} for fast rendering and memory efficiency. 
$\beta_i$ is a learnable parameter to control the individual Gaussian density distribution and effective region. In our SolidGS, we use a global shared solidness factor $\beta_g$ for all Gaussians. This way, we can better control the solidity of all Gaussians and make them grow solid during the training. 

As training proceeds, \(\beta_g\) automatically grows larger. As shown in Fig.~\ref{fig:gef} (a), the distribution approximates the uniform distribution and the Gaussian consolidates between \( \left( \bmu - \bsigma, \bmu + \bsigma \right) \). In \cref{fig:gef} (b), when \( \beta_g = 20\), the Gaussian already has unified alpha blending weight \(\alpha_i\) over most of its effective area. The distance rendered would be identical regardless of the view directions on the Gaussian. This consistent geometry rendering would stabilize the optimization and consequently promote the final surface reconstruction.

\subsection{Optimizations}

\label{sec:optimization}

3D Gaussian splatting \cite{kerbl20233d} employs an RGB reconstruction loss, $\cL_c$, combining $\cL_1$ and D-SSIM terms between rendered and ground truth images, to encourage realistic image synthesis. 
However, with sparse-view inputs, this loss is not sufficient to form a consistent 3D geometry. To address this, we introduce additional geometric regularization to better guide Gaussian rotations and positions. We also recognized the limited constraints that sparse views impose on surface quality. We incorporate monocular normal priors from visual foundation models to enhance geometric accuracy. We put all optimization terms in \cref{fig:flow}.

\subsubsection{Geometric Regularizations}
\setlength\tabcolsep{0.5em}
\begin{table*}[t]
\vspace{-2pt}
\centering

\caption{\textbf{Quantitative Comparison on DTU Dataset~\cite{jensen2014large}.} We show the Chamfer Distance. Our SolidGS achieves the highest reconstruction accuracy among other methods. We also mark the best, second-best, and third-best results in red, orange, and yellow respectively.}
\vspace{-2mm}
\resizebox{.9\textwidth}{!}{

\begin{tabular}{@{}l|ccccccccccccccclc}
\toprule
Method  & 24   & 37   & 40   & 55   & 63   & 65   & 69   & 83   & 97   & 105  & 106  & 110  & 114  & 118  & 122  &  & Mean  \\
\midrule
PGSR \cite{chen2024pgsr}      
& 2.37 & 2.67 & 2.72 & 1.35 & 1.62 & 2.91 & \tbest 1.14 & 1.43 & 1.78 & 1.09 & 2.30 & 2.14 & 0.80 & 1.56 & 1.61  & & 1.83 \\
2DGS \cite{huang20242d}      
& 2.05 & \tbest 2.57 & \tbest 2.25 & 1.33 & 1.74 & 2.29 & 1.39 & 1.46 & 1.52 & 1.06 & 1.92 & 1.87 & 0.84 & 1.73 & 1.59  & & 1.71 \\
\midrule
DNGaussian \cite{li2024dngaussian} 
& 3.48 & 5.71 & 3.69 & 3.13 & 3.72 & \tbest 2.20 & 4.23 & 2.92 & 4.59 & 3.65 & 6.79 & 5.30 & 2.97 & 5.44 & 3.62  & & 4.10 \\
MVPGS \cite{xu2024mvpgs}     
& \tbest 1.93 & \best 2.31 & \best 1.99 & 1.07 & 1.44 & 2.97 & 1.31 &  1.37 & \sbest 1.23 & \best 1.04 & \tbest 1.85 & \best 1.02 & 0.83 & 1.54 & 1.33  & & \tbest 1.55 \\
\midrule
SparseNeus \cite{long2022sparseneus}  
& 4.81 & 5.56 & 5.81 & 2.68 & 3.30 & 3.88 & 2.39 & 2.91 & 3.08 & 2.33 & 2.64 & 3.12 & 1.74 & 3.55 & 2.31  & & 3.34 \\
NeuSurf \cite{huang2024neusurf}   
& \best 1.35 & 3.25 & 2.50 & \sbest 0.80 & \sbest 1.21 & 2.35 & \best 0.77 & \best 1.19 & \best 1.20 & \sbest 1.05 & \best 1.05 & \tbest 1.21 & \best 0.41 & \best 0.80 & \sbest 1.08  & & \sbest 1.35 \\
SparseCraft \cite{younes2024sparsecraft}   
& 2.29 & 2.67 & 2.99 & \best 0.69 & \tbest 1.43 & \sbest 2.18 & 1.16 & \tbest 1.30 & 1.52 & 1.14 & 1.86 & - & \sbest 0.56 & \tbest 1.00 & \tbest 1.12 & & 1.57 \\
\midrule
Ours       
& \sbest 1.50 & \sbest 2.40 & \sbest 2.20 & \tbest 0.98 & \best 0.88 & \best 1.36 & \sbest 0.79 & \sbest1.29 & \tbest 1.33 & \tbest 1.06 & \sbest 1.72 & \sbest 1.07 & \tbest 0.60 & \sbest 0.99 & \best 0.89  & & \best 1.27 \\
\bottomrule
\end{tabular}
}
\vspace{-2mm}
\label{tab:dtu_result}
\end{table*}
\label{sec:regularization}

\boldparagraph{Normal consistency loss} 
Following \cite{huang20242d, yu2024gaussian, chen2024pgsr}, the normal consistency loss $\cL_{nc}$ measures the consistency between the directly rendered normal and the normal calculated from the depth map,

\begin{equation}
    \cL_{nc}=\sum_i \alpha_i \left(1-\bn_i^{\top} \bN_i \right),
\end{equation}

\noindent where \(i\) is the index of the pixel, and \(\bn_i\) represents the normal calculated from the gradient of the depth map. This loss ensures local smoothness of the rendered depth. 

\boldparagraph{Depth distortion loss}
We also use the depth distortion loss $\cL_d$ in \cite{huang20242d, yu2024gaussian}, which penalizes variations in depth among Gaussians on the same ray and reduces depth inconsistency,
\begin{equation}
    \cL_d=\sum_{i, j} \alpha_i \alpha_j \left|\bD(p)_i - \bD(p)_j\right|, 
\end{equation}

\noindent where \(\alpha_i = \rho_i \cG_i(x)\) is the blending weight.

\boldparagraph{Constraints on virtual views}
To avoid overfitting during sparse view training, we generate unseen virtual views and then apply the aforementioned regularization \( \cL_{nc} \) and \( \cL_d \) to these views. The virtual views are randomly generated as linear interpolations of a pair of training views' poses, added with small perturbations. This way we have better consistency and smoother reconstructions of the shared area among training views.

\subsubsection{Geometric Priors}
\label{sec:priors}
To emphasize the geometric correctness of Gaussian splitting, we use the off-the-shelf monocular normal predictor, Metric3D \cite{hu2024metric3d}, to provide additional geometric clues. Given the three input views, we feed them into Metric3D and get corresponding predicted normals \(\hat{\bN}\). Similar to~\cite{dai2024high}, we directly apply L1 loss and cosine loss between rendered normal and predicted normal:
\vspace{-2mm}
\begin{equation}
    \cL_{nr} =  \sum_i \left(\left|\bN_i - \hat{\bN_i}\right| + \left(1 - \bN_i \cdot \hat{\bN_i} \right) \right).
\end{equation}
\vspace{-2mm}

Additionally, for the smoothness of the depth map, we apply the same normal loss to normal calculated depth map:
\vspace{-2mm}
\begin{equation}
    \cL_{nd} =  \sum_i \left(\left|\bn_i - \hat{\bN_i}\right| + \left(1 - \bn_i \cdot \hat{\bN_i} \right) \right).
\end{equation}
\vspace{-2mm}

\subsection{Training}
\boldparagraph{Initialization}
Due to the sparsity of training input, a dense initialization can help speed up optimization. Therefore, we follow \cite{xu2024mvpgs, yao2018mvsnet} and leverage the MVSFormer \cite{cao2022mvsformer} to generate 3D point clouds as initial positions of the Gaussians. 

\boldparagraph{Loss function}
The loss that we use to regularize the virtual views is:
\vspace{-1mm}
\begin{equation}
    \cL_{virtual} = \lambda_d \cL_d + \lambda_{nc} \cL_{nc},
\end{equation}

\noindent and the total loss to train the input views is:
\vspace{-1mm}
\begin{equation}
    \cL = \cL_{virtual} + \cL_c + \lambda_{nr} \cL_{nr} + \lambda_{nd} \cL_{nd} + \lambda_1 \cL_s,
\end{equation}

\noindent where \(\cL_s\) is the flatten 3D Gaussian Loss used in PGSR \cite{chen2024pgsr} to form flattened Gaussians, and \( \lambda_d = 10000 \), \( \lambda_{nc} = \lambda_{nr} = \lambda_{nd} = 0.015 \), and \( \lambda_1 = 100\) are preset constant weights. Since monocular normal estimation does not guarantee global consistency across the frame, we set the weights of two monocular normal guided losses \(\lambda_{nr} \) and \( \lambda_{nd} \) the same as the normal consistency loss \( \lambda_{nc} \), hoping them to work at the same strength as regularization term and the RGB loss still dominants the training process.

Notice that we don't put any constraints on the global solidness factor $\beta_g$ during training. However, $\beta_g$ tends to converge to maximize itself under our optimization, resulting in the Gaussian getting more solid. 

\boldparagraph{Mesh extraction}
We render the depth maps from the input views and then fuse them into a TSDF \cite{curless1996volumetric}. Then we extract the surface mesh using Marching Cube \cite{lorensen1998marching}. With highly consistent depth, we can extract a high-quality mesh with details from only 3 depth maps.
\begin{table}[t]
  \centering
  \caption{\textbf{Performance Comparison on DTU Dataset~\cite{jensen2014large}.} PSNR is missing from SparseNeus~\cite{long2022sparseneus} and NeuSurf~\cite{huang2024neusurf} since they only reconstruct colorless meshes. Best results are highlighted in \textbf{bold}. 
  }

\vspace{-2mm}
\resizebox{.9\linewidth}{!}{
  \begin{tabular}{@{}l|ccc}
    \toprule
Method  & Mean CD~$\downarrow$  & PSNR~$\uparrow$  & Training Time~$\downarrow$             \\
    \midrule
PGSR~\cite{chen2024pgsr}      & 1.83 & 20.80 & 2.2min                 \\
2DGS~\cite{huang20242d}       & 1.71 & 20.65 & \textbf{2.1min}                 \\
DNGaussian~\cite{li2024dngaussian} & 4.10 & 18.91 & 3.8min                 \\
MVPGS~\cite{xu2024mvpgs}      & 1.55 & 20.24 & 2.3min                 \\
SparseNeus~\cite{long2022sparseneus} & 3.34 & -     & \textgreater 24h \\
NeuSurf~\cite{huang2024neusurf}    & 1.35 & -     & \textgreater 8h            \\
SparseCraft~\cite{younes2024sparsecraft}    & 1.57 & 20.55    & 10 min            \\
Ours       & \textbf{1.27} & \textbf{21.32} & 3.1min                 \\
    \bottomrule
\end{tabular}
}
\vspace{-2mm}
  \label{tab:dtu_nvs}
\end{table}

\begin{table}[t]
  \centering
  \caption{\textbf{Quantitative Comparison on TNT Dataset~\cite{knapitsch2017tanks}.} The best results are highlighted in \textbf{bold}.}

\vspace{-2mm}
\resizebox{.9\linewidth}{!}{
  \begin{tabular}{@{}l|cccc|c}
    \toprule
Method & Barn & Truck & Caterpillar & Ignatius & Avg  \\
    \midrule
PGSR~\cite{chen2024pgsr}  & 0.06 & 0.09  & 0.02        & 0.13     & 0.08 \\
2DGS~\cite{huang20242d}   & 0.20 & 0.27  & 0.12        & 0.25     & 0.21 \\
MVPGS~\cite{xu2024mvpgs}  & \textbf{0.26} & 0.27  & \textbf{0.17}        & 0.31     & 0.25 \\
Ours   & \textbf{0.26} & \textbf{0.29}  & 0.14        & \textbf{0.33}     & \textbf{0.26} \\
    \bottomrule
\end{tabular}
}
\vspace{-5mm}
  \label{tab:tnt_precision}
\end{table}


\begin{figure*}[t]
  \centering
   \includegraphics[width=0.9\linewidth]{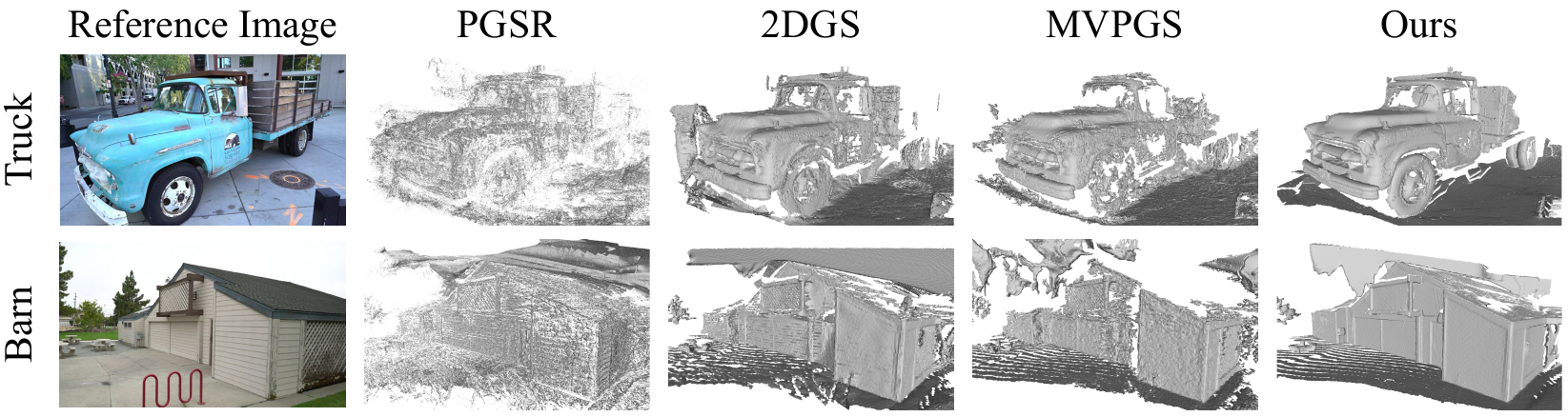}
   \vspace{-2mm}
   \caption{\textbf{Qualitative Mesh Results on TNT Dataset.} All meshes are extracted using TSDF + Marching Cube from Gaussians.}
   \label{fig:tnt_mesh}
   \vspace{-2mm}
\end{figure*}

\begin{figure*}[t]
  \centering
   \includegraphics[width=0.9\linewidth]{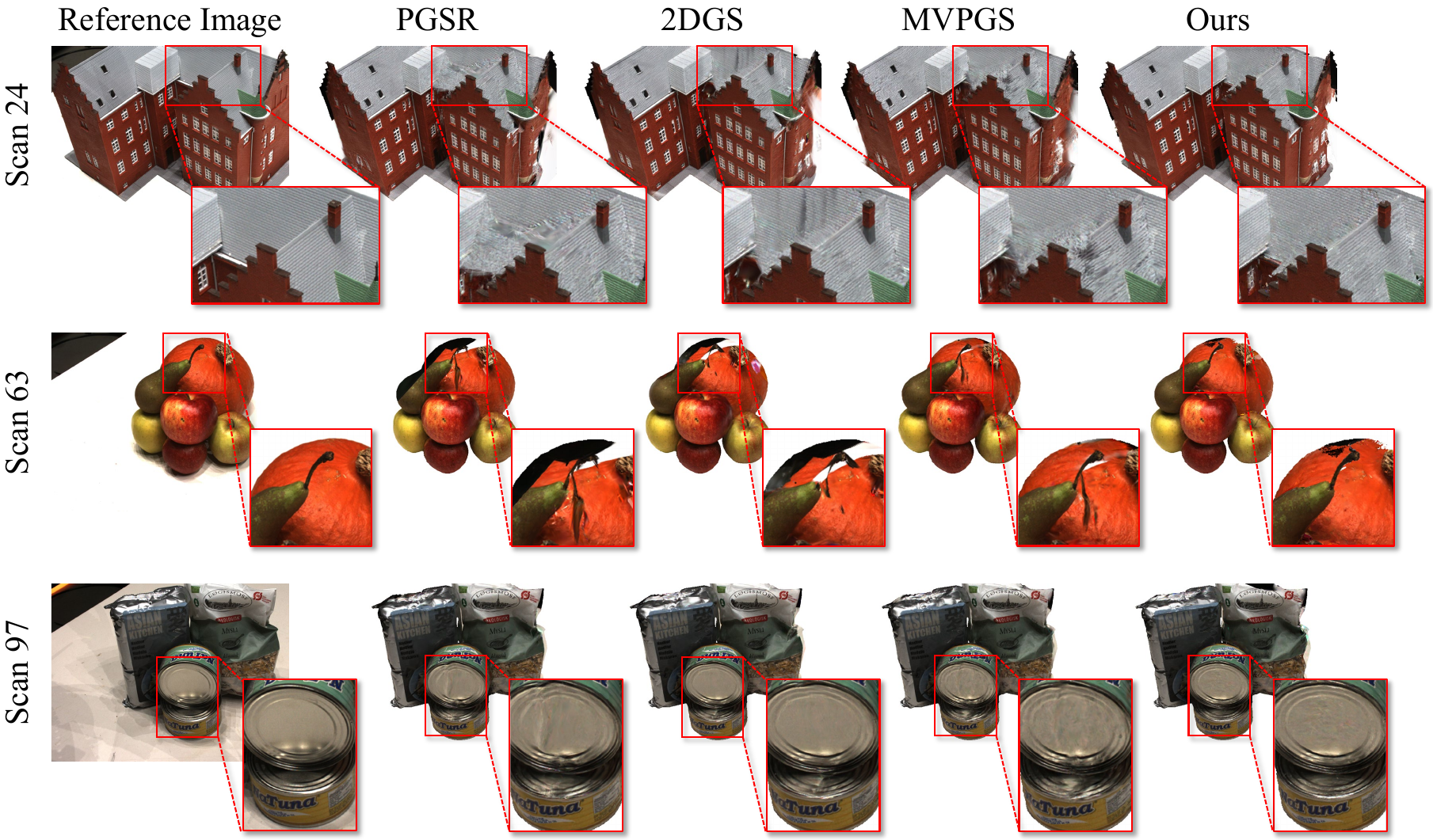}
   \caption{\textbf{Qualitative NVS Results on DTU Dataset.} We demonstrate novel view synthesis results of our method against PGSR~\cite{chen2024pgsr}, 2DGS~\cite{li2024dngaussian}, and MVPGS~\cite{xu2024mvpgs}. Backgrounds are masked out in all inference images.
   }
   \label{fig:dtu_nvs}
\end{figure*}

\section{Experiments}

\subsection{Experiment Setup}

\boldparagraph{Datasets}
We conduct our experiments on three real-world datasets, DTU \cite{jensen2014large}, Tanks and Temples(TNT)~\cite{knapitsch2017tanks}, and LLFF \cite{mildenhall2019local}. DTU dataset consists of 15 scenes of a single object with RGB images and depth scans. Foreground masks are used during evaluation following \cite{niemeyer2022regnerf} and mesh extraction. TNT dataset contains large-scale indoor and outdoor scenes with intricate geometry. LLFF dataset contains RGB images of forward-facing scenes. We split the DTU and LLFF datasets following \cite{niemeyer2022regnerf, li2024dngaussian, xu2024mvpgs, huang2024neusurf} and split TNT with similar viewing angles variance in LLFF to train our method on 3 input images and test from unseen views. 


\boldparagraph{Metrics}
In DTU, we evaluate our mesh results by Chamfer Distance (CD) between reconstructed meshes and the fused ground-truth depth scans. We also report the PSNR in terms of NVS quality. In TNT, we report the precision of the reconstructed mesh. In LLFF, we visualize the mesh for visual comparison.

\boldparagraph{Baselines}
We compare our SolidGS with representative reconstruction methods, which can be mainly categorized into three groups: 1. Gaussian-based explicit dense-view surface reconstruction methods, such as PGSR \cite{chen2024pgsr} and 2DGS \cite{huang20242d}; 2. Gaussian-based explicit sparse-view NVS methods, such as DNGaussian \cite{li2024dngaussian} and MVPGS \cite{xu2024mvpgs}; and 3. neural implicit sparse-view surface reconstruction methods, such as SparseNeus \cite{long2022sparseneus}, NeuSurf \cite{huang2024neusurf}, and SparseCraft~\cite{younes2024sparsecraft}. For a fair comparison, we use the same point clouds from NVSFormer \cite{cao2022mvsformer} as initialization in PGSR and 2DGS. 

\boldparagraph{Implementation Details}
Our method is built upon the public code of PGSR \cite{chen2024pgsr} with our modified CUDA kernels so that it can take in shared solidity factor and other geometric regularization. We incorporate default training parameters and the densification strategy in 3DGS, except that we change the total iterations to 10000 and the densification stops at 5000 iterations. Geometric regularization is involved in the training starting at 1000 iterations. During training, we reset the solidity factor every 1000 iterations in the first 5000 iterations. Virtual cameras are used every 50 iterations. We conducted all our experiments on a desktop with an i9-13900K CPU and an RTX 4090 GPU.

\subsection{Surface Reconstruction Results}

In \cref{tab:dtu_result} and \cref{tab:dtu_nvs}, we compare our surface reconstruction to the baselines we selected on Chamfer Distance and training time on the DTU dataset. Our method achieves the highest geometry reconstruction quality with the least CD among all compared methods. Compared to the dense view methods 2DGS~\cite{huang20242d} and PGSR~\cite{chen2024pgsr}, our method significantly enhances the surface quality in each scan with little training time overhead. Compared to implicit methods SparseNeus~\cite{long2022sparseneus},  NeuSurf~\cite{huang2024neusurf}, and SparseCraft~\cite{younes2024sparsecraft}, our method benefits from the fast rendering of Gaussian Splatting and trains significantly faster with better geometry. 

\cref{fig:dtu_mesh} provides the qualitative comparison of the reconstructed mesh. Among all explicit methods, our mesh achieves the highest quality in terms of geometry accuracy, completeness, and detail sharpness. Compared to the PGSR and 2DGS, our method captures better details due to the geometry priors. Due to our SolidGS, the depths are precise and consistent across views, which gives fewer floating artifacts and faithful geometry to ground truth. MVPGS renders highly inconsistent depth maps and reconstructs broken meshes. The Chamfer Distance doesn't reflect the mesh completeness, therefore, MVPGS have the lowest CD in some scenes. Mesh completeness is more obviously demonstrated in \cref{fig:dtu_mesh}.
In comparison with the neural implicit method NeuSurf, our method reconstructs the sharper edges and cutoffs between objects. 

We demonstrate our method's capability in reconstructing large-scale scenes from sparse input in \cref{tab:tnt_precision} and \cref{fig:tnt_mesh}. PGSR~\cite{chen2024pgsr} cannot converge to a consistent geometry across multiple input views, resulting in a broken mesh and low precision scores. Our method reaches the highest average precision compared to 2DGS~\cite{huang20242d} and MVPGS~\cite{xu2024mvpgs}. Our method is superior in mesh completeness and correctness notably from the visual results.

More results including the LLFF dataset can be found in the Supplementary.

\subsection{Novel View Synthesis Results}

Our approach models scenes as a radiance field, enabling the rendering of high-quality images from new perspectives. We compare our Novel View Synthesis (NVS) results against all baselines in \cref{tab:dtu_result}. With the correct geometry, our method has fewer artifacts in the final Gaussian field and reaches the highest PSNR in the NVS task. A qualitative comparison is performed in \cref{fig:dtu_nvs}. For the roof in scan 24, the pumpkin in scan 63, and the cans in scan 97, there all exist visual artifacts due to geometry mismatches in the baselines. Our method reconstructs geometry that is plausible from novel views, greatly improving the NVS results.

\subsection{Ablations}
\begin{table}[t]
  \centering
  \vspace{-4mm}
  \caption{\textbf{Quantitative Ablation Study on DTU dataset~\cite{jensen2014large}.}}
  \resizebox{.96\linewidth}{!}{
  \begin{tabular}{@{}l|ccc}
    \toprule
Setting              & Accuracy~$\downarrow$ & Completion ~$\downarrow$ & Average~$\downarrow$ \\
    \midrule
Base                  & 1.76 & 1.91 & 1.83 \\
Base + Regularization & 1.50 & 1.76 & 1.63 \\
\tbest Base + SolidGS        & \tbest 1.43 & \tbest 1.72 & \tbest 1.58 \\
\sbest Base + Mono Normal    & \sbest 1.18 & \sbest 1.54 & \sbest 1.36 \\
\best Full                  & \best 1.09 & \best 1.45 & \best 1.27 \\
    \bottomrule
  \end{tabular}
  }
  \vspace{-2mm}
  \label{tab:ablation}
\end{table}

\vspace{+2mm}
\begin{figure}[t]
  \centering
   \includegraphics[width=\linewidth]{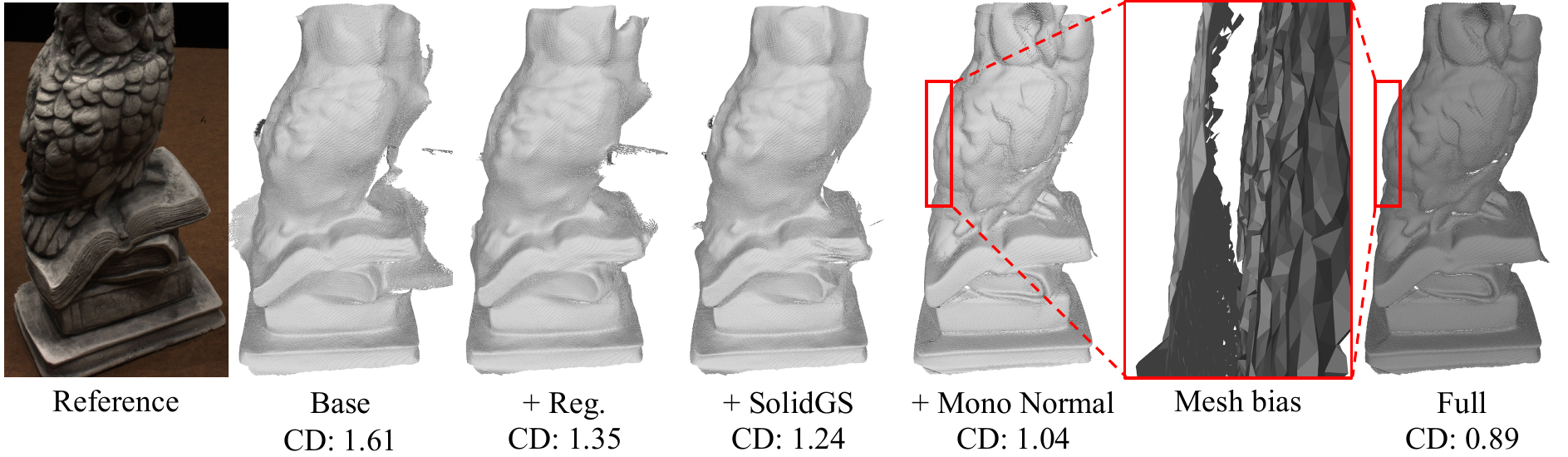}
   \caption{\textbf{Qualitative ablation on DTU.} We simultaneously visualize the mesh of "+ Mono Normal" and "Full" inside the red box.
   }
\vspace{-1mm}
   \label{fig:ablation}
\vspace{-3mm}
\end{figure}

We conduct our ablation study on the DTU dataset (\cref{tab:ablation} and \cref{fig:ablation}). Starting from our baseline PGSR \cite{chen2024pgsr}, we first show the effect of additional geometric regularizations, including virtual camera and depth distortion loss. This modification gives us a smoother geometry over limited training views, which is reflected by a less erroneous shape on the fused mesh. Then we show the effect of our modified SolidGS. Compared to the based model, this modification makes Gaussian closer to solid surfels during training, thus leading to a more accurate depth blending and fewer floater artifacts in the fused mesh. Third, by introducing monocular normal prior, we have a mesh that is self-consistent in shape. Notice that when we compare this modification against the full model, there is a huge bias between the two meshes, which leads to a higher Chamfer distance. That's because our SolidGS gives high accuracy and uniform geometry during training, and therefore mitigates the depth inconsistency caused by the inaccurate geometry rendering. Putting every component together, our method reconstructs a smooth, precise, and complete mesh.  

\subsection{Limitations and Future Works}

\begin{figure}
  \centering
  \begin{subfigure}{0.6\linewidth}
    \centering
    \includegraphics[width=0.95\linewidth]{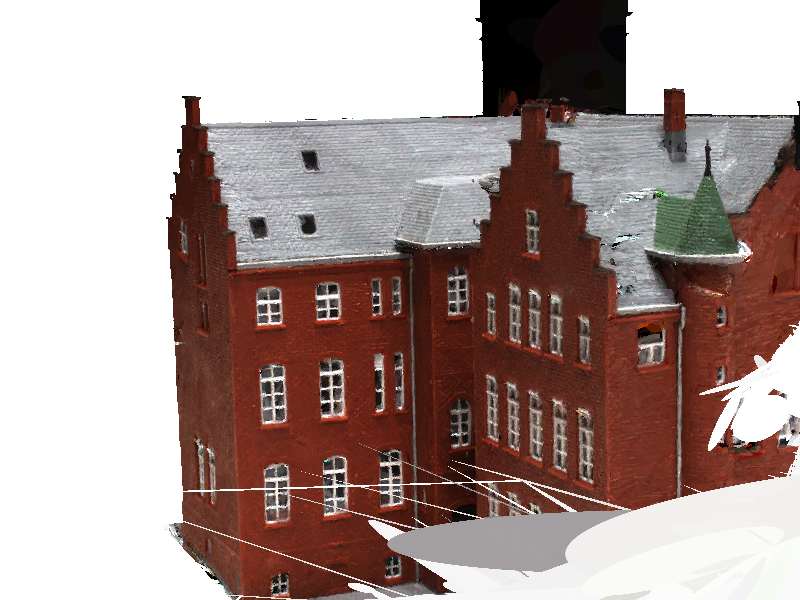}
    \caption{Renderings with large viewpoint change. }
    \label{fig:limitation-b}
  \end{subfigure}
  \hfill
  \begin{subfigure}{0.35\linewidth}
    \centering
    \includegraphics[width=0.95\linewidth]{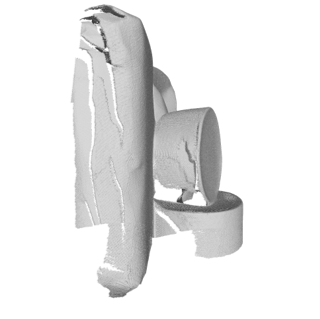}
    \caption{Cracks exist on regions visible to one view. }
    \label{fig:limitation-a}
  \end{subfigure}
\vspace{-3mm}
  \caption{\textbf{Limitations.} Artifacts occur with novel views.}
   \vspace{-5mm}
  \label{fig:limitation}
\end{figure}

Although our method achieves great progress in extracting a mesh from overlapping areas from sparse view inputs, our method can be unstable in areas that appear in only one training view. This is reflected by the cracks on the edge of the mesh, as shown in \cref{fig:limitation-a} and the rendering quality could largely decrease with the deviation of rendering viewpoints from the input sparse view as shown in \cref{fig:limitation-b}. To mitigate this limitation, it is worth investigating identifying local geometry connectivity from training views and enforcing this connectivity during training. We leave all the above exploration for future works.


\section{Conclusion}
In conclusion, we present SolidGS, a method for enhancing sparse-view surface reconstruction through consolidated Gaussian splatting. By introducing SolidGS with a global learnable solidness factor, our approach promotes multi-view geometry consistency, leading to high-fidelity reconstructions even with limited input views. The integration of geometric regularizations and priors further supports stable optimization, allowing SolidGS to produce detailed and precise meshes within minutes. Experiments show that our methods achieve state-of-the-art in both surface reconstruction and novel view synthesis from sparse inputs.


{
    \small
    \bibliographystyle{ieeenat_fullname}
    \bibliography{main}
}

\clearpage
\setcounter{page}{1}

\maketitlesupplementary

%
\section{Summary}

This supplementary includes a \textbf{video} demonstrating the qualitative results of geometry reconstruction and Novel View Synthesis. We also showcase more results on DTU, Tanks-and-Temples, and LLFF datasets. We compare the effect of training view number. We also include a discussion on the concurrent works.

\section{More results}

\subsection{Results on Tanks-and-Temples Dataset}

\setlength\tabcolsep{0.5em}
\begin{table*}[t]
\vspace{-2pt}
\centering

\caption{\textbf{Quantitative Comparison on DTU Dataset~\cite{jensen2014large} with Different Input Views}. We show the Accuracy (Accu.), Completion (Comp), and Chamfer Distance (CD). We mark the best and second-best results in red and orange respectively.}
\vspace{-2mm}
\resizebox{.98\textwidth}{!}{
\begin{tabular}{@{}l|ccc|ccc|ccc|ccc|ccc}
\toprule
\multicolumn{1}{c}{\multirow{2}{*}{Methods}} & \multicolumn{3}{c}{2 view}   & \multicolumn{3}{c}{3 view}   & \multicolumn{3}{c}{4 view}   & \multicolumn{3}{c}{5 view}   & \multicolumn{3}{c}{6 view}   \\
\cline{2-16} 
\multicolumn{1}{c}{}                         & Accu. $\downarrow$ & Comp. $\downarrow$ & CD $\downarrow$  & Accu. $\downarrow$ & Comp. $\downarrow$ & CD $\downarrow$  & Accu. $\downarrow$ & Comp. $\downarrow$ & CD $\downarrow$  & Accu. $\downarrow$ & Comp. $\downarrow$ & CD $\downarrow$  & Accu. $\downarrow$ & Comp. $\downarrow$ & CD $\downarrow$   \\
\midrule
\midrule
PGSR                                         & 2.02     & 2.50       & 2.26 & 1.76     & 1.91       & 1.83 & 1.46     & 1.55       & 1.51 & 1.10     & 1.26       & 1.18 & 0.84     & 0.98       & 0.91 \\
2DGS                                         & 1.84     & 2.16       & 2.00 & \sbest 1.54     & 1.88       & 1.71 & 1.45     & 1.74       & 1.59 & 1.32     & 1.64       & 1.48 & 1.10     & 1.46       & 1.28 \\
\midrule
MVPGS                                        & \best 1.02     &  \sbest 2.87       &  \sbest 1.94 & \best 1.09     &  \sbest 2.01       &  \sbest 1.55 & \best 0.95     & \sbest 1.55       & \sbest 1.25 & \sbest 0.81     & \sbest 1.35       & \sbest 1.08 & \sbest 0.73     & \sbest 1.15       & \sbest 0.94 \\
\midrule
Ours                                         & \sbest 1.27     & \best 1.95       & \best 1.61 &  \best 1.09     & \best 1.45       & \best 1.27 &  \sbest 1.01     & \best 1.26       & \best 1.13 & \best 0.80     & \best 0.97       & \best 0.89 & \best 0.66     & \best 0.83       & \best 0.75 \\
\bottomrule
\end{tabular}
}
\vspace{-2mm}
\label{tab:views}
\end{table*}

\begin{figure*}[t]
  \centering
   \includegraphics[width=0.9\linewidth]{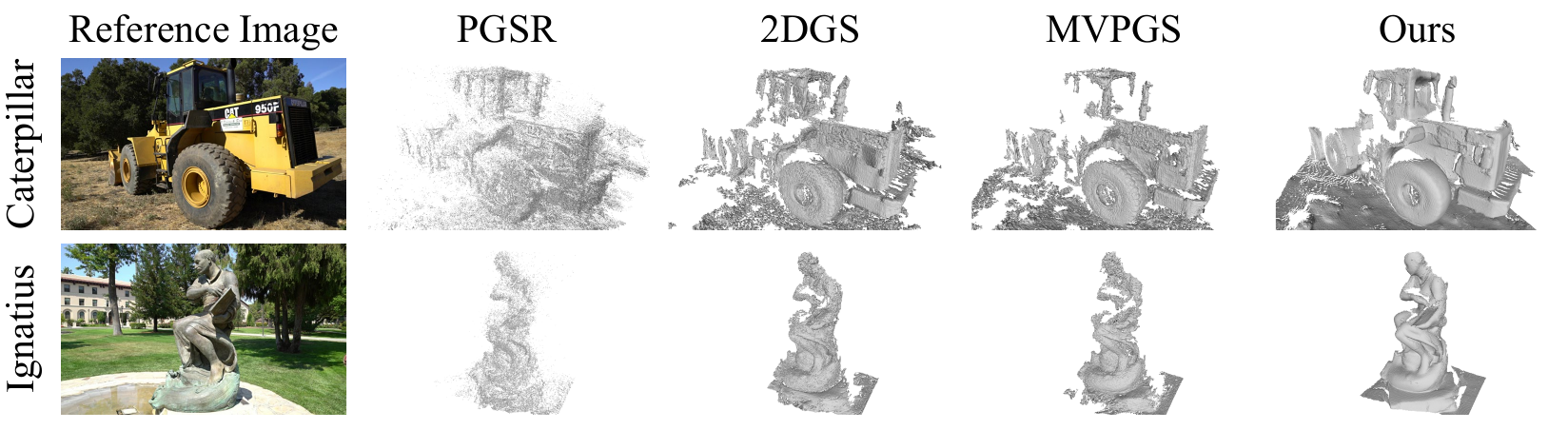}
   \caption{\textbf{Qualitative Mesh Results on TNT Dataset~\cite{knapitsch2017tanks}.} All meshes are extracted using TSDF + Marching Cube from Gaussians.}
   \label{fig:tnt_mesh_supp}
\end{figure*}

In \cref{fig:tnt_mesh_supp} we provide more qualitative comparisons of our method against all explicit reconstruction techniques on the Tanks-and-Temples (TNT) dataset~\cite{knapitsch2017tanks}. In the Caterpillar and Ignatius scenes, our method reconstructs surfaces with higher completeness of the centroid structure.

\subsection{Results on LLFF Dataset}

\begin{figure*}[t]
  \centering
   \includegraphics[width=0.9\linewidth]{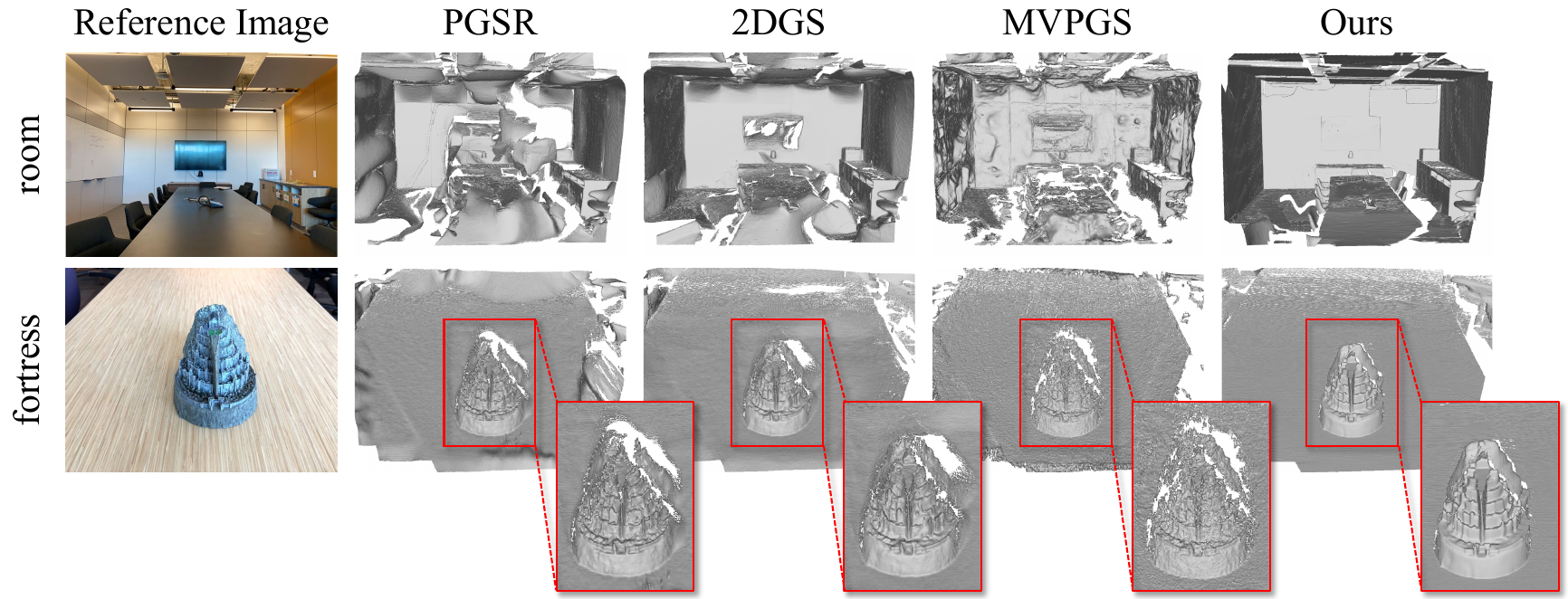}
   \caption{\textbf{Qualitative Mesh Results on LLFF Dataset~\cite{mildenhall2019local}.} All meshes are extracted using TSDF + Marching Cube from Gaussians.}
   \label{fig:llff_mesh}
\end{figure*}

In \cref{fig:llff_mesh} we present a qualitative comparison of our method against all explicit reconstruction techniques on the LLFF dataset~\cite{mildenhall2019local}. In the Room scene, our approach effectively reconstructs the walls and tables, whereas other methods exhibit distorted geometry and contain holes. In the Fortress scene, our method accurately captures the shape of the fortress with reduced noise on the mesh. Additionally, our method produces a flatter and more complete table in the reconstructed surface. 

\subsection{Different Input Views}

\cref{tab:views} presents the quantitative results for various input views on the DTU Dataset. Our method consistently achieves the highest Completion and Chamfer Distance across all settings, indicating its robustness with different training views. Additionally, as the number of training views increases, our method's accuracy improves more rapidly compared to MVPGS. This highlights its greater potential from additional training views.

\begin{figure*}[t]
  \centering
   \includegraphics[width=0.9\linewidth]{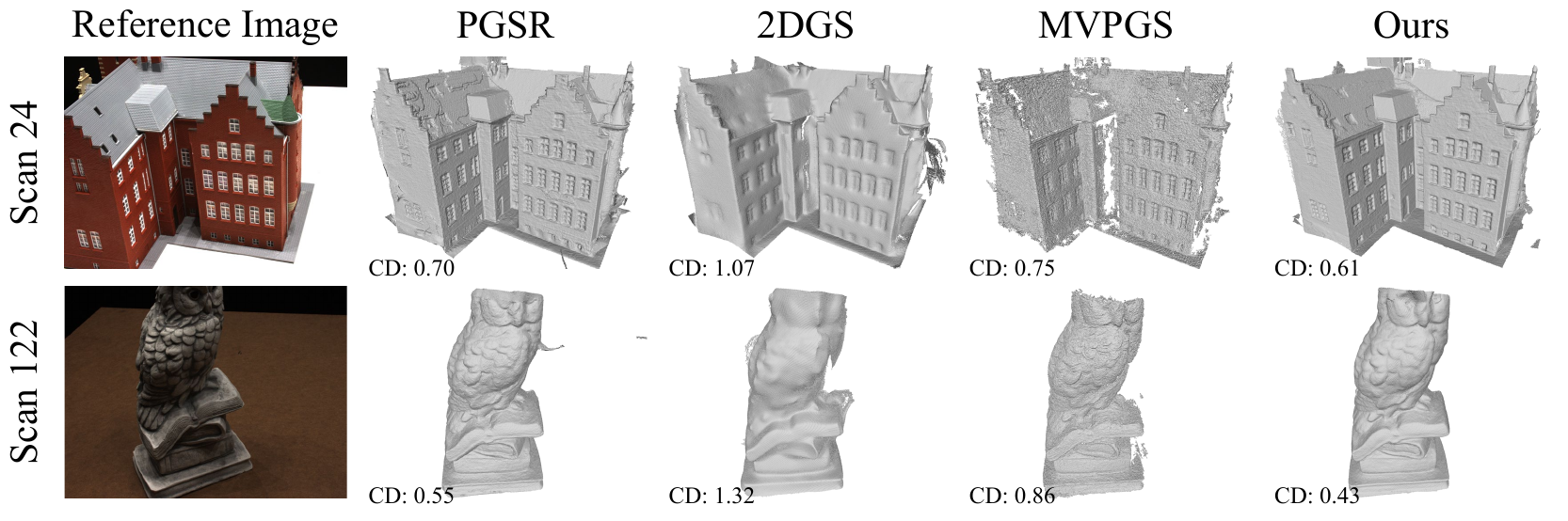}
   \caption{\textbf{Qualitative Mesh Results on DTU Dataset~\cite{jensen2014large} with 6 Input Views.}}
   \label{fig:dtu_mesh_6view}
\end{figure*}

\begin{figure*}[t]
  \centering
   \includegraphics[width=0.9\linewidth]{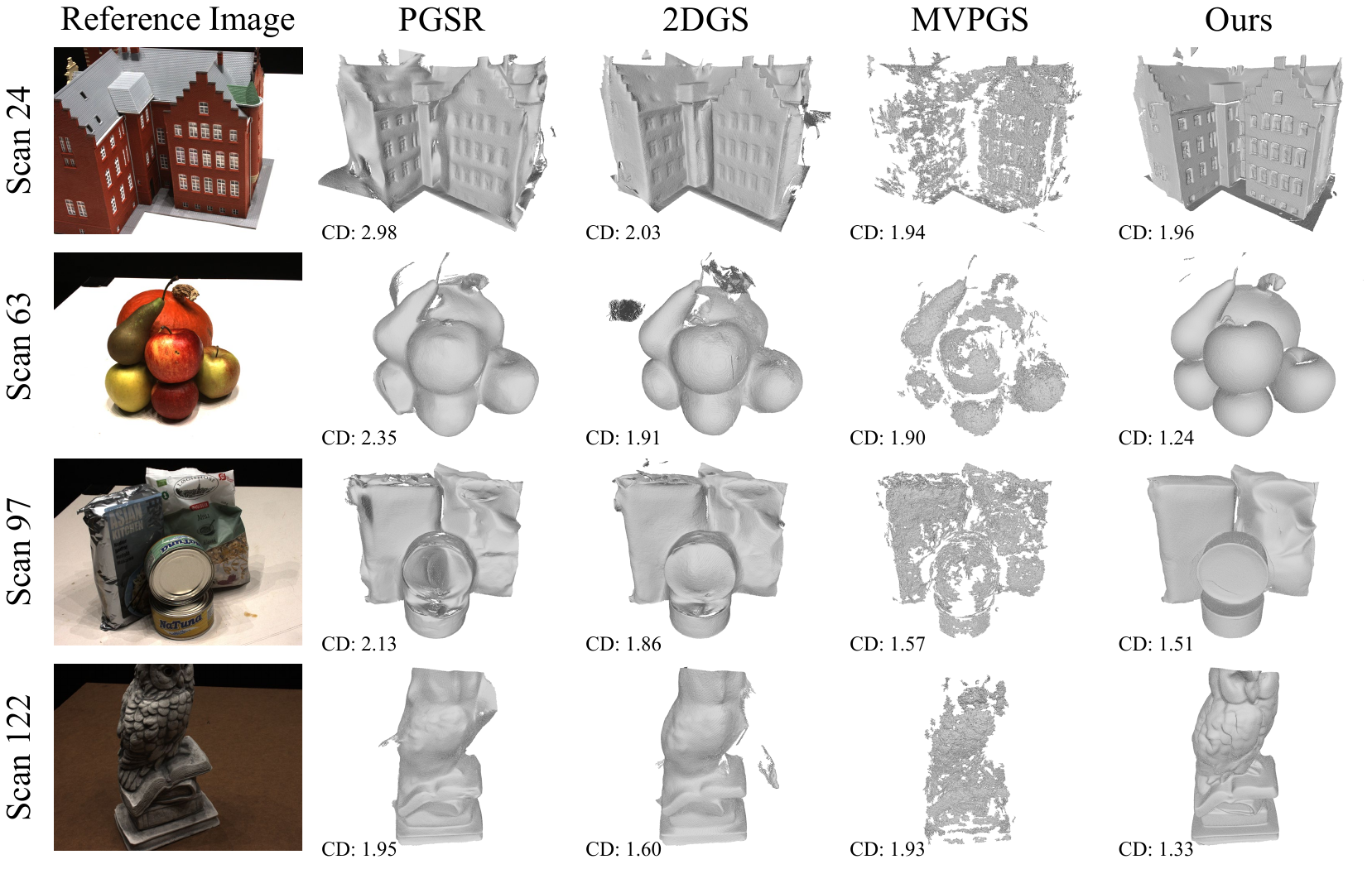}
   \caption{\textbf{Qualitative Mesh Results on DTU Dataset~\cite{jensen2014large} with 2 Input Views.}}
   \label{fig:dtu_mesh_2view}
\end{figure*}

\cref{fig:dtu_mesh_6view} displays our qualitative results with six input views, demonstrating that increased input views enhance reconstruction quality with finer details.
We also present our qualitative results of reconstructed mesh in \cref{fig:dtu_mesh_2view}. Even with extremely sparse input (2 views), our method still reconstructs plausible results, surpassing all other methods. 

\section{Discussion on Concurrent Works}
We observe that concurrent works such as SpikingGS~\cite{zhang2024spiking} and 2DGH~\cite{yu20242dgh} have modified Gaussian kernel functions in Gaussian splatting to enhance geometric reconstruction accuracy. SpikingGS~\cite{zhang2024spiking} employs spiking neurons on the opacity of Gaussians and their kernel functions, effectively eliminating semi-transparent Gaussians and their tails. Although our motivations align, our approach, SolidGS, reconstructs a more continuous volume representation, resulting in superior novel view quality. Conversely, 2DGH~\cite{yu20242dgh} replaces the traditional Gaussian kernel with a Gaussian-Hermite kernel to augment the expressiveness of individual Gaussians. However, this method introduces increased ambiguity, particularly in under-supervised regions. Our SolidGS focuses on consolidating opacity during rendering, thereby mitigating multiview geometry inconsistencies. Additionally, SolidGS incorporates more geometric constraints to achieve high-fidelity surface reconstruction, even under challenging sparse input conditions.


\end{document}